\documentclass[preprint,12pt]{elsarticle}
\usepackage{epsfig}
\usepackage{amssymb}
\usepackage{amsmath}
\usepackage{amsthm}
\usepackage{caption}
\usepackage{subcaption}
\usepackage{mathabx}
\newcommand{\ignore}[1]{}
\usepackage{lipsum}
\makeatletter
\def\ps@pprintTitle{%
 \let\@oddhead\@empty
 \let\@evenhead\@empty
 \def\@oddfoot{}%
 \let \@oddfoot}
\makeatother
\begin{document}

\begin{frontmatter}

\title{On Data-Driven Saak Transform}

\author{C.-C. Jay Kuo and Yueru Chen}
\address{Ming-Hsieh Department of Electrical Engineering \\
University of Southern California, Los Angeles, CA 90089-2564, USA}

\begin{abstract}

Being motivated by the multilayer RECOS (REctified-COrrelations on a
Sphere) transform, we develop a data-driven Saak ({\bf S}ubspace {\bf
a}pproximation with {\bf a}ugmented {\bf k}ernels) transform in this
work. The Saak transform consists of three steps: 1) building the
optimal linear subspace approximation with orthonormal bases using the
second-order statistics of input vectors, 2) augmenting each transform
kernel with its negative, 3) applying the rectified linear unit (ReLU)
to the transform output. The Karhunen-Lo$\acute{e}$ve transform (KLT) is
used in the first step.  The integration of Steps 2 and 3 is powerful
since they resolve the sign confusion problem, remove the rectification
loss and allow a straightforward implementation of the inverse Saak
transform at the same time.  Multiple Saak transforms are cascaded to
transform images of a larger size.  All Saak transform kernels are
derived from the second-order statistics of input random vectors in a
one-pass feedforward manner.  Neither data labels nor backpropagation is
used in kernel determination.  Multi-stage Saak transforms offer a
family of joint spatial-spectral representations between two extremes;
namely, the full spatial-domain representation and the full
spectral-domain representation.  We select Saak coefficients of higher
discriminant power to form a feature vector for pattern recognition, and
use the MNIST dataset classification problem as an illustrative example. 

\end{abstract}

\end{frontmatter}

\section{Introduction}\label{sec:introduction}

Signal transforms provide a way to convert signals from one
representation to another. For example, the Fourier transform maps a
time-domain function into a set of Fourier coefficients. The latter
representation indicates the projection of the time-domain function onto
a set of orthonormal sinusoidal basis functions. The orthonormal basis
facilitates the inverse transform. The original function can be
synthesized by summing up all Fourier basis functions weighted by their
Fourier coefficients. The basis functions (or transform kernels) are
typically selected by humans. One exception is the
Karhunen-Lo$\acute{e}$ve transform (KLT) \cite{stark1986probability}.
The KLT kernels are the unit eigenvectors of the covariance matrix of
sampled data. It is the optimal transform in terms of energy compaction.
That is, to obtain an approximation to an input signal class, we can
truncate part of KLT basis functions associated with the smallest
eigenvalues.  The truncated KLT provides the optimal approximation to
the input with the smallest mean-squared-error (MSE). 

We develop new data-driven forward and inverse transforms in this work.
For a set of images of size $N \times N$, the total number of variables
in these images is $N^2$ and their covariance matrix is of dimension
$N^4$. It is not practical to conduct the KLT on the full image for a
large $N$. Instead, we may decompose images into smaller blocks and
conduct the KLT on each block. To give an example, the Discrete Cosine
Transform (DCT) \cite{ahmed1974discrete} provides a good approximation
to the KLT for image data, and the block DCT is widely used in the
image/video compression standards. One question of interest is whether
it is possible to generalize the KLT so that it can be applied to images
of a larger size in a hierarchical fashion? Our second research
motivation comes from the resurgent interest on convolutional neural
networks (CNNs) \cite{Nature2015,juang2016deep}.  The superior
performance of CNNs has been demonstrated in many applications such as
image classification, detection and processing. To offer an explanation,
Kuo \cite{kuo2016understanding, kuo2017cnn} modeled the convolutional
operation at each CNN layer with the RECOS (REctified-COrrelations on a
Sphere) transform, and interpreted the whole CNN as a multi-layer RECOS
transform. 

By following this line of thought, it would be a meaningful task to
define the inverse RECOS transform and analyze its properties.  The
analysis of the forward/inverse RECOS transform will be conducted in
Sec.  \ref{sec:RECOS}.  Being similar to the forward and inverse Fourier
transforms, the forward and inverse RECOS transforms offer tools for
signal analysis and synthesis, respectively.  However, unlike the
data-independent Fourier transform, the RECOS transform is derived from
labeled training data, and its transform kernels (or filter weights) are
optimized by backpropagation.  The analysis of forward/inverse RECOS
transforms is challenging due to nonlinearity.  We will show that the
RECOS transform has two loss terms: the approximation loss and the
rectification loss.  The approximation loss is caused by the use of a
limited number of transform kernels. This error can be reduced by
increasing the number of filters at the cost of higher computational
complexity and higher storage memory.  The rectification loss is due to
nonlinear activation.  Furthermore, since the filters in the RECOS
transform are not orthogonal to each other, the inverse RECOS transform
demands the solution of a linear system of equations. 

It is stimulating to develop a new data-driven transform that has
neither approximation loss nor the rectification loss as the RECOS
transform.  Besides, it has a set of orthonormal transform kernels so
that its inverse transform can be performend in a straightforward
manner.  To achieve these objectives, we propose the Saak ({\bf
S}ubspace {\bf a}pproximation with {\bf a}ugmented {\bf k}ernels)
transform.  As indicated by its name, the Saak transform has two main
ingredients: 1) subspace approximation and 2) kernel augmentation.  To
seek the optimal subspace approximation to a set of random vectors, we
analyze their second-order statistics and select orthonormal
eigenvectors of the covariance matrix as transform kernels. This is the
well-known KLT. When the dimension of the input space is very large
(say, in the order of thousands or millions), it is difficult to conduct
one-stage KLT. Then, we may decompose a high-dimensional vector into
multiple lower-dimensional sub-vectors.  This process can be repeated
recursively to form a hierarchical representation. For example, we can
decompose one image into four non-overlapping quadrants recursively to
build a quad-tree whose leaf node is a small patch of size $2 \times 2$.
Then, a KLT can be defined at each level of the quad-tree. 

If two or more transforms are cascaded directly, there is a ``sign
confusion" problem \cite{kuo2016understanding, kuo2017cnn}.  To resolve
it, we insert the Rectified Linear Unit (ReLU) activation function in
between.  The ReLU inevitably brings up the rectification loss, and a
novel idea called kernel augmentation is proposed to eliminate this
loss.  That is, we augment each transform kernel with its negative
vector, and use both original and augmented kernels in the Saak
transform.  When an input vector is projected onto the positive/negative
kernel pair, one will go through the ReLU operation while the other will
be blocked.  This scheme greatly simplifies the signal representation
problem in face of ReLU nonlinear activation.  It also facilitates the
inverse Saak transform.  The integration of kernel augmentation and ReLU
is equivalent to the sign-to-position (S/P) format conversion of the
Saak transform outputs, which are called the Saak coefficients.  By
converting the KLT to the Saak transform stage by stage, we can cascade
multiple Saak transforms to transform images of a large size.  The
multi-stage Saak transforms offer a family of joint spatial-spectral
representations between two extremes - the full spatial-domain
representation and the full spectral-domain representation. The Saak and
multi-stage Saak transforms will be elaborated in Sec. \ref{sec:Saak}
and Sec.  \ref{sec:M-Saak}, respectively. 

Although both CNNs and multi-stage Saak transforms adopt the ReLU, it is
important to emphasize one fundamental difference in their filter
weights (or transform kernels) determination. CNN's filter weights are
determined by the training data and their associated labels.  After
initialization, these weights are updated via backpropagation.  A CNN
determines its optimal filter weights by optimizing a cost function via
backpropagation iteratively.  The iteration number is typically huge.
The multi-stage Saak transforms adopt another fully automatic method in
determining their transform kernels. They are selected based on the
second-order statistics of input vectors at each stage.  It is a
one-pass feedforward process from the leaf to the root. Neither data
labels nor backpropagation is needed for transform kernel determination. 

The Saak coefficients in intermediate stages indicate the spectral
component values in the corresponding covered spatial regions. The Saak
coefficients in the last stage represent spectral component values of
the entire input vector (i.e., the whole image).  We use the MNIST
dataset as an example to illustrate the distribution of Saak
coefficients of each object class. Based on the ANalysis Of VAriance
(ANOVA), we select Saak coefficients of higher discriminant power by
computing their F-test score. The larger the F-test score, the higher
the discriminant power.  Finally, we compare the classification accuracy
with the support vector machine (SVM) and the K-Nearest-Neighbors (KNN)
classifiers. 

The rest of this paper is organized as follows. The forward and inverse
RECOS transforms are studied in Sec. \ref{sec:RECOS}.  The forward and
inverse Saak transforms are proposed in Sec. \ref{sec:Saak}.  The
multi-stage Saak transforms are presented in Sec. \ref{sec:M-Saak}.  The
application of multi-stage Saak transforms to image classification for
the MNIST dataset is described in Sec. \ref{sec:applications}.  The CNN
approach and the Saak-transform-based machine learning methodology are
compared in Sec. \ref{sec:discussion}. Finally, concluding remarks are
given and future research directions are pointed out in Sec.
\ref{sec:conclusion}. 

\section{RECOS Transform}\label{sec:RECOS}

The RECOS transform is a data-driven transform proposed in
\cite{kuo2016understanding, kuo2017cnn} to model the convolutional
operations in a CNN. In the context of image processing, the forward
RECOS transform defines a mapping from a real-valued function defined on
a three-dimensional (3D) cuboid to a one-dimensional (1D) rectified
spectral vector. The forward and inverse RECOS transforms will be
studied in this section. 

\begin{figure}[htb]
\centering
\includegraphics[width=10cm]{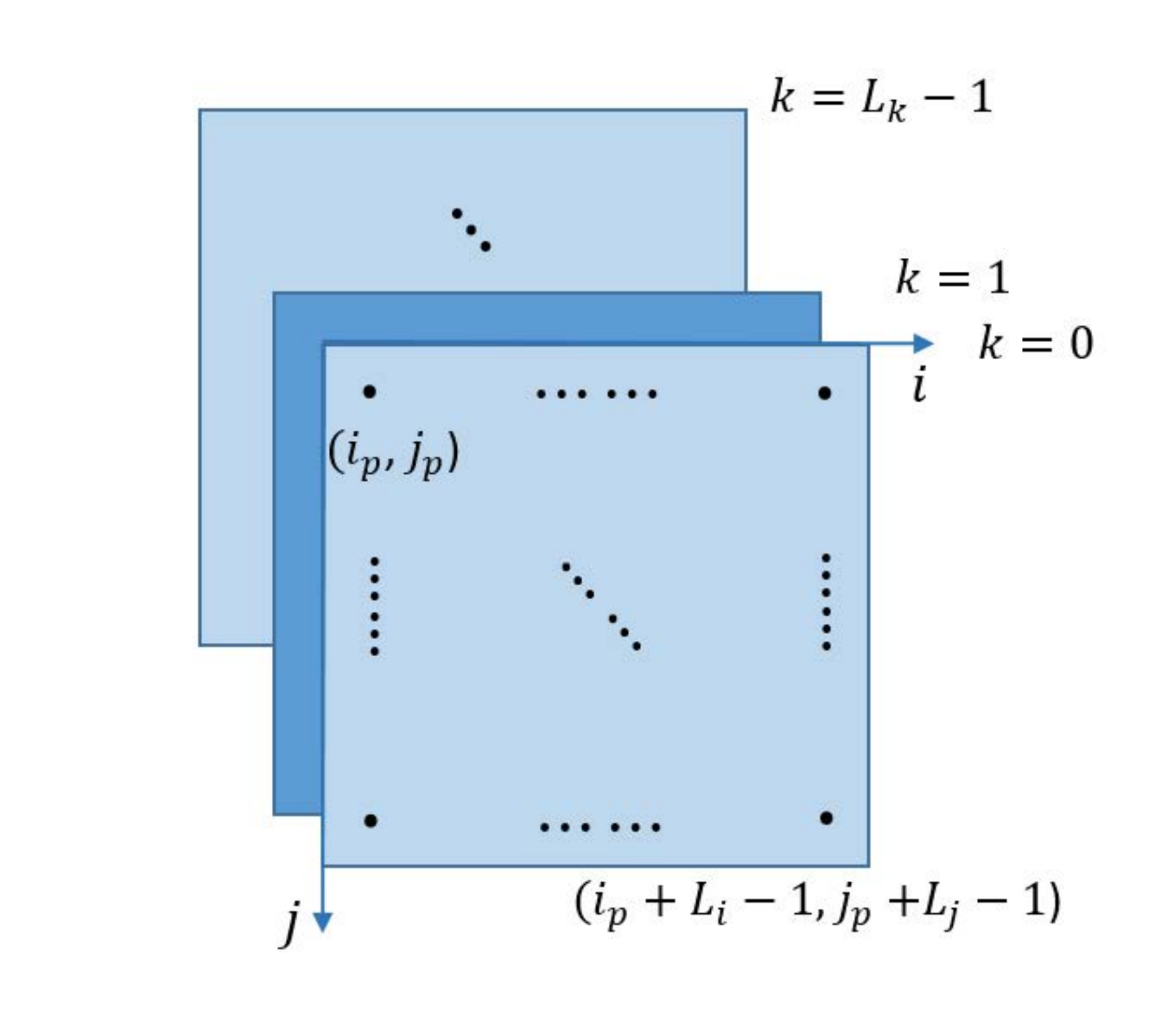}
\caption{Illustration of a cuboid with its pivot at $(i,j)=(i_p,j_p)$.}\label{fig:ssc}
\end{figure}

\subsection{Forward RECOS Transform}\label{subsec:f-RECOS}

As illustrated in Fig. \ref{fig:ssc}, a spatial-spectral cuboid, denoted
by $C(i_p, j_p, L_i, L_j, L_k)$, consists of 3D grid points with indices
$(i,j,k)$
\begin{itemize}
\item along the horizontal dimension: $i \in \{i_p, i_p+1, \cdots, i_p+L_i-1 \}$,
\item along the vertical dimension:  $j \in \{j_p, j_p+1, \cdots, j_p+L_j-1   \}$,
\item along the spectral dimension: $k \in \{0, 1, 2, \cdots L_k-1\}$,
\end{itemize}
where $(i,j)=(i_p,j_p)$ is its spatial pivot and $L_i$, $L_j$ and $L_k$
are its width, height and depth, respectively.  For a real-valued
function defined on cuboid $C(i_p, j_p, L_i, L_j,L_k)$, one can flatten
these values into a one-dimensional (1D) vector,
\begin{equation}\label{eqn:input}
{\bf f} \in R^N, \quad \mbox{where} \quad N=L_i \times L_j \times L_k,
\end{equation}
by scanning the 3D grid points with a fixed order. All vectors in
$R^N$ are generated by the same flattening rule. 

Consider an anchor vector set \cite{kuo2016understanding, kuo2017cnn}
that contains $L_k$ vectors of unit length,
\begin{equation}\label{eqn:anchorset}
A = \{ {\bf a}_0, {\bf a}_1, \cdots, {\bf a}_k, \cdots, {\bf a}_K \}, 
\quad ||{\bf a}_k||=1 \mbox{ and } K=L_k-1,
\end{equation}
where ${\bf a}_k$ is defined on  $C(i_p, j_p, L_i, L_j,L_k)$ and flattened 
to a vector in $R^N$. We divide anchor vectors into two types. The vector
\begin{equation}\label{eqn:DC}
{\bf a}_0=\frac{1}{\sqrt{N}} (1, 1, \cdots, 1)^T
\end{equation}
is the DC anchor vector while the remaining ones, ${\bf a}_1, \cdots,
{\bf a}_K$, are the AC anchor vectors.  An effective RECOS transform
depends on careful selection of AC anchor vectors. A common way to
select anchor vectors is to train the CNN with labeled data via
backpropagation.  Given anchor vector set $A$ in Eq.
(\ref{eqn:anchorset}), the forward RECOS transform can be summarized 
as follows. 

\noindent
{\bf Forward RECOS Transform.} 
The forward RECOS transform consists of two steps.
\begin{enumerate}
\item Project input ${\bf f} \in R^N$ onto $K+1$ anchor vectors to
get $K+1$ projection values\footnote{The input ${\bf f}$ was assumed to
be a mean-removed vector of unit length in \cite{kuo2016understanding,
kuo2017cnn}. This assumption is no more needed here.}:
\begin{equation}\label{eqn:projection}
p_k={\bf a}_k^T {\bf f}, \quad k=0, 1, \cdots K.
\end{equation}
Since anchor vector ${\bf a}_k$ plays the transform kernel role as
shown in Eq. (\ref{eqn:projection}), the two terms ``anchor vectors"
and ``transform kernels" are used interchangeably. We call 
\begin{equation}\label{eqn:p-vector}
{\bf p}=(p_0, p_1, \cdots, p_K)^T \in R^{K+1}
\end{equation}
the projection vector of input ${\bf f}$. 
\item Preserve the DC projection (i.e., $p_0$) and pass AC projections
(i.e., $p_1, \cdots, p_K$) through the ReLU activation function.  
The output ${\bf g} \in R^{K+1}$ of the RECOS transform can be 
written as
\begin{equation}\label{eqn:output}
{\bf g}=(g_0, g_1, \cdots, g_K)^T , 
\end{equation}
where 
\begin{equation}
g_0=p_0,
\end{equation}
and
\begin{equation}\label{eqn:AC_output}
g_k= \left\{
\begin{tabular}{ll}
$p_k$, & \quad if \quad $p_k > 0$, \\
$0$, & \quad if \quad $p_k \le 0$.
\end{tabular}
\right. k=1, 2, \cdots, K.
\end{equation}
\end{enumerate}
In above, we allow $g_0$ to be either positive or negative while $g_k$,
$k=1,\cdots,K$, is non-negative. The DC projection contributes to the
bias term. The forward RECOS transform defined above offers a
mathematical description of convolutional operations in both
convolutional and fully connection layers of CNNs. 

\subsection{Inverse RECOS Transform}\label{subsec:i-RECOS}

The inverse RECOS transform reconstructs input ${\bf f} \in R^N$ as
closely as possible from its output ${\bf g} \in R^{K+1}$ of the forward
RECOS transform.  To proceed, we first examine the reconstruction of the
input from the unrectified projection vector, ${\bf p}$, as given in Eq.
(\ref{eqn:p-vector}).  When $K+1 < N$, we cannot reconstruct input ${\bf
f} \in R^N$ exactly but its approximation in the linear space spanned by
vectors in $A$, i.e.  \begin{equation}\label{eqn:reconstruction-1} 
{\bf f} \approx \hat{\bf f} = \sum_{k=0}^{K} {\alpha}_k {\bf a}_k.
\end{equation}
Because of Eqs. (\ref{eqn:projection}) and (\ref{eqn:reconstruction-1}), 
we get
$$
p_k \approx {\bf a}_k^T \hat{\bf f} = {\bf a}_k^T \left( \sum_{k'=0}^{K} 
{\alpha}_{k'} {\bf a}_{k'} \right), \quad k=0, 1, \cdots, K.
$$
If the approximation error $||\hat{\bf f}-{\bf f}||$ is negligible, we 
can set up a linear system of equations in form of
$$
p_k= \sum_{k'=0}^{K} {\bf a}_k^T {\bf a}_{k'} {\alpha}_{k'}, \quad k=0, 1, 
\cdots, K,
$$
with unknowns ${\alpha}_{k}$. If the $(K+1)\times (K+1)$ coefficient 
matrix has the full rank, we can solve for ${\alpha}_{k}$ uniquely.

\noindent
{\bf Inverse RECOS Transform.} All AC outputs of the RECOS transform are
either positive or zero as given in Eq. (\ref{eqn:AC_output}).  We can
use a permutation matrix to re-arrange $g_1, \cdots, g_K$ so that all
zero elements are moved to the end of the re-ordered vector. Suppose
that there are $1 \le Q \le K$ nonzero elements in $g_1, \cdots, g_K$.
After re-arrangement, we can express it as
\begin{equation}\label{re-arranged}
{\bf g}' = ( g_0, g'_1, \cdots, g'_Q,  0, \cdots, 0 ) ^T,
\end{equation}
which has $K-Q$ zero elements at the end. The re-arrangement of $g_1,
\cdots, g_K$ is the same as re-arranging AC anchor vectors, ${\bf a}_1,
\cdots, {\bf a}_K$. We use ${\bf a'}_1, \cdots, {\bf a'}_Q$ to denote
anchor vectors associated with projections $p'_1, \cdots, p'_Q$. Then,
we obtain
\begin{equation}\label{eqn:reconstruction-2}
{\bf f}' = \sum_{q=0}^{Q} \beta_q {\bf a'}_q, 
\end{equation}
where ${\bf a'}_0={\bf a}_0$.  Eqs. (\ref{eqn:reconstruction-1}) and
(\ref{eqn:reconstruction-2}) are the reconstruction formulas of ${\bf f}
\in R^N$ using unrectified and rectified projection vectors,
respectively. 

The inverse RECOS transform is a mapping from a $(Q+1)$-dimensional
rectified vector $(p_0, p'_1, \cdots, p'_Q)^T$ to an approximation of
${\bf f}$ in the $(Q+1)$-dimensional subspace as defined in Eq.
(\ref{eqn:reconstruction-2}). By following the same procedure given
above, we have
$$
p_q \approx {\bf a}_q^T \left( \sum_{q'=0}^{Q} {\beta}_{q'} {\bf a'}_{q'} 
\right), \quad q=0, 1, \cdots, Q,
$$
due to Eq. (\ref{eqn:projection}), Eq. (\ref{eqn:reconstruction-2}) and
${\bf f} \approx {\bf f}'$.  Besides the approximation loss, there is an
additional loss caused by the rectification of the ReLU unit in form of
\begin{equation}\label{rectification_error} 
E_r ( \hat{\bf f}, {\bf f}') = || \hat{\bf f} - {\bf f}' ||^2,
\end{equation} 
which is called the rectification loss. 

Let $\Psi$, $\hat{\Psi}$ and $\Psi'$ denote the space of ${\bf f} \in R^N$, the
linear space spanned by $A$ and the linear space spanned by
\begin{equation}\label{eqn:Q-anchorset} 
A' = \{ {\bf a}_0, {\bf a}'_1, \cdots, {\bf a}'_Q \}. 
\end{equation}
It is clear from Eqs. (\ref{eqn:reconstruction-1}) and 
(\ref{eqn:reconstruction-2}) that 
$$
\Psi' \subset \hat{\Psi} \subset \Psi.
$$
The loss between ${\bf f}$ and ${\bf f}'$ can be computed as
\begin{equation}\label{eqn:ineq}
E ( {\bf f}, {\bf f}') =  || {\bf f} - {\bf f}' ||^2 =
|| {\bf f} - \hat{\bf f} ||^2 + || \hat{\bf f} - {\bf f}' ||^2 
+ 2 ({\bf f} - \hat{\bf f})^T (\hat{\bf f} - {\bf f}').
\end{equation}
Recall that $|| {\bf f} - \hat{\bf f} ||^2 = E_a ( {\bf f}, \hat{\bf
f})$ is the approximation loss and $|| \hat{\bf f} - {\bf f}' ||^2 = E_r
( \hat{\bf f}, {\bf f}')$ is the rectification loss.  The term $2 ({\bf
f} - \hat{\bf f})^T (\hat{\bf f} - {\bf f}')$ becomes zero if $({\bf f}
- \hat{\bf f})$ and $(\hat{\bf f} - {\bf f}')$ are orthogonal to each
other. 

If $|| {\bf f} - {\bf f}' ||^2$ is negligible, we can set up a linear
system of $(Q+1)$ equations in form of
$$
p_q = \sum_{q'=0}^{Q} {\bf a}_q^T {\bf a}_{q'} {\alpha}_{q'} \quad q=0, 1, 
\cdots, Q,
$$
with $(Q+1)$ unknowns ${\beta}_{q}$, $q=0, 1, \cdots, Q$. Again, if the
$(Q+1)\times (Q+1)$ coefficient matrix is of full rank, we can solve
${\alpha}_{q'}$ uniquely. It is worthwhile to point out that the inverse
RECOS transform is well defined once all anchor vectors are specified. 

\section{Saak Transform}\label{sec:Saak}

\subsection{Forward Saak Transform} 

The forward/inverse RECOS transforms have several limitations.
\begin{enumerate}
\item It is desired to have orthonormal transform kernels to facilitate
forward and inverse transforms. In other words, anchor vectors should
satisfy the following condition:
\begin{equation}\label{eqn:delta}
{\bf a}_i^T {\bf a}_j = < {\bf a}_i,{\bf a}_j >  = \delta_{i,j},
\end{equation} 
where $0 \le i,j \le K$ and $\delta_{i,j}$ is the Kroneckor delta
function.  Under this condition, the reconstruction formulas in Eqs.
(\ref{eqn:reconstruction-1}) and (\ref{eqn:reconstruction-2}) can be
greatly simplified as
\begin{equation}\label{eqn:reconstruction-1a}
\hat{\bf f} = \sum_{k=0}^{K} p_k {\bf a}_k, \quad \mbox{and} \quad
{\bf f}' = \sum_{q=0}^{Q} p_q {\bf a'}_q, 
\end{equation}
respectively. Furthermore, the loss between ${\bf f}$ and ${\bf f}'$ in 
Eq. (\ref{eqn:ineq}) can be computed exactly as
\begin{equation}\label{eqn:ineq-2}
E({\bf f},{\bf f}')=E_a({\bf f},\hat{\bf f})+E_r(\hat{\bf f},{\bf f}'),
\end{equation}
since $({\bf f} - \hat{\bf f})$ and $(\hat{\bf f} - {\bf f}')$ are
orthogonal to each other. Unfortunately, anchor vectors in the RECOS
transform do not satisfy the condition in Eq. (\ref{eqn:delta}).
\item Once the network architecture is fixed, the RECOS transform cannot
make the approximation loss, $E_a({\bf f},\hat{\bf f})$, arbitrarily 
small. The approximation error has a certain floor.
\item The rectification operation makes signal representation less
efficient since the representation power of some anchor vectors is lost.
It is desired to recover the representation capability of these rectified
anchor vectors. 
\end{enumerate}

To address the first two problems, we consider the
Karhunen-Lo$\acute{e}$ve transform (KLT) on a set of mean-removed input
vectors.  Since the projection of input ${\bf f} \in R^N$ on the DC
anchor vector gives its mean, the residual
\begin{equation}\label{eqn:residual}
\tilde{\bf f}={\bf f}-p_0 {\bf a}_0
\end{equation}
is a zero-mean random vector. We compute the correlation matrix of
$\tilde{\bf f}$, ${\bf R}=E[\tilde{\bf f} \tilde{\bf f}^T] \in R^{N
\times N}$. Matrix ${\bf R}$ has $(N-1)$ positive eigenvalues and one
zero eigenvalue. The eigenvector associated with the zero eigenvalue is
the constant vector. The remaining $(N-1)$ unit-length eigenvectors
define the Karhunen-Lo$\acute{e}$ve (KL) basis functions, which are
KLT's kernel vectors. It is well known that the KLT basis functions are
orthonormal. Besides, we can order them according to their eigenvalues
from the largest to the smallest, and the first $M$ (with $M \le N$)
basis functions provide the optimal linear approximation subspace $R^M$
in $R^N$. Consequently, we can choose these $M$ orthonormal kernels to
yield the minimum approximation error. The above-mentioned procedure is
usually known as the Principal Component Analysis (PCA) or the truncated
KLT. 

\begin{figure}[htb]
\centering
\includegraphics[width=8cm]{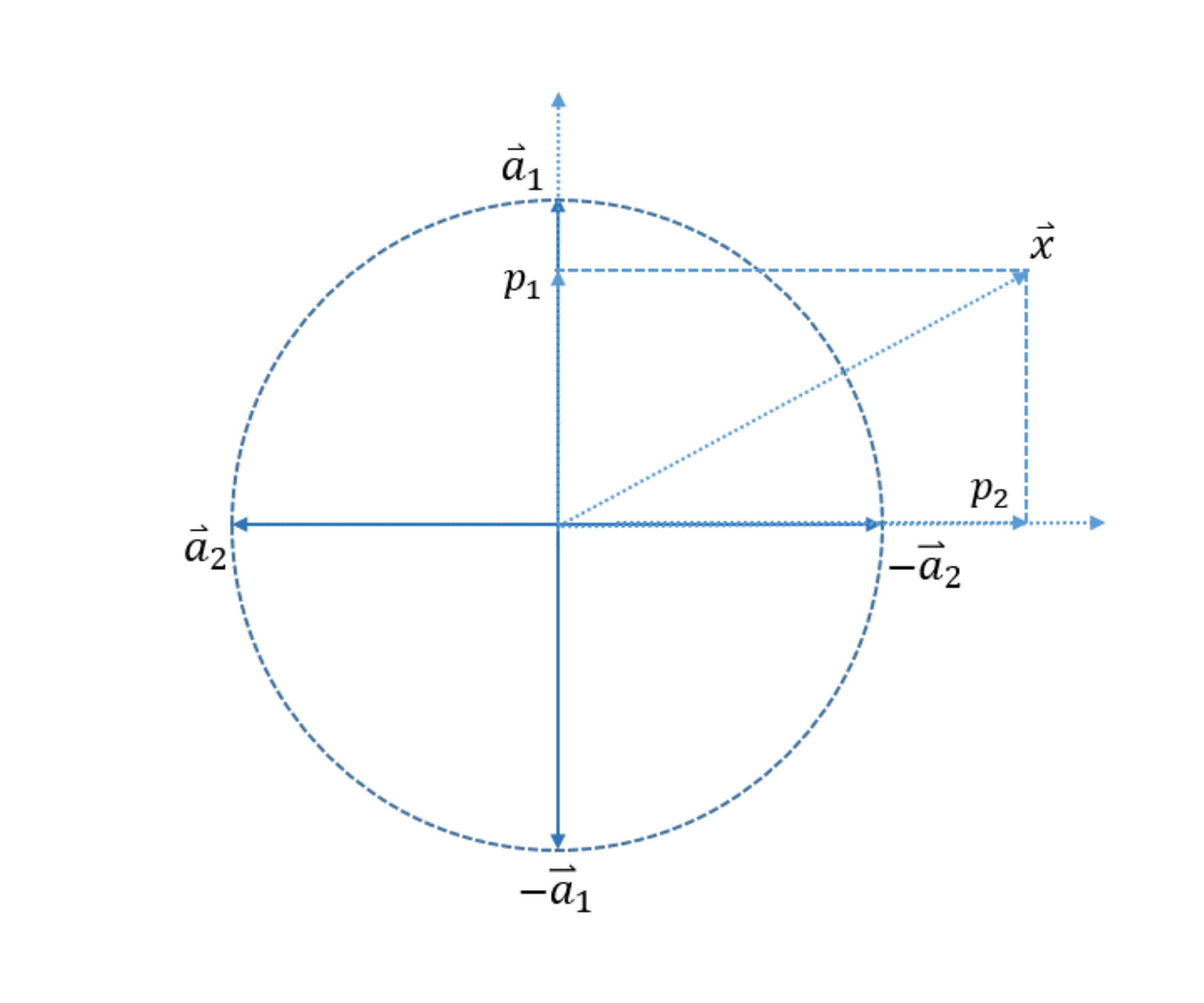} 
\caption{Illustration of the kernel augmentation idea.}
\label{fig:aug_components}
\end{figure}

To address the third problem, we propose to augment each kernel vector
with its negative vector. That is, if ${\bf a}_k$ is a kernel, we choose
$-{\bf a}_k$ to be another kernel. An example is shown in Fig.
\ref{fig:aug_components}. In this figure, we show the projection of
input ${\bf f}$ onto two AC anchor vectors ${\bf a}_1$ and ${\bf a}_2$
with $p_1 > 0$ and $p_2 <0$ as their respective projection values.  The
ReLU preserves $p_1$ but clips $p_2$ to 0. By augmenting them with two
more anchor vectors, $-{\bf a}_1$ and $-{\bf a}_2$, we obtain projection
values $-p_1 < 0$ and $-p_2 >0$.  The ReLU clips $-p_1$ to 0 and
preserves $-p_2$. The kernel augmentation idea is powerful in two
aspects: 1) eliminating the rectification loss, and 2) annihilating the
nonlinear effect of ReLU to greatly simplify the analysis.  By following
the notations in Sec.  \ref{sec:RECOS}, we summarize the forward Saak
transforms below. \vspace{6ex}

\noindent
{\bf Forward Saak Transform.} 
\begin{enumerate}
\item Kernel Selection and Augmentation \\
Collect a representative set of input samples ${\bf f} \in
R^N$ and determine its KLT basis functions, which are denoted by
${\bf b}_k$, $k=1, \cdots, N$. The DC kernel ${\bf a}_0$ is
defined by Eq. (\ref{eqn:DC}). The remaining $2(N-1)$ AC kernels
are obtained using the augmentation rule:
\begin{equation}\label{eqn:projection-2}
{\bf a}_{2k-1}={\bf b}_k, \quad  {\bf a}_{2k}=-{\bf b}_k, \quad k=1, \cdots, N-1.
\end{equation}
\item Projection onto the Augmented Kernel Set \\
Project input ${\bf f}$ on the augmented kernel set from Step 1:
\begin{equation}\label{eqn:project-2}
p_k= {\bf a}_k^T {\bf f},
\end{equation}
and 
\begin{equation}\label{eqn:p-vector-2}
{\bf p}=(p_0, p_1, \cdots, p_{2(N-1)})^T \in R^{2N-1}
\end{equation}
is the projection vector of input ${\bf f}$. 
\item Apply the ReLU to the projection vector except the first element 
to yield the final output:
\begin{equation}\label{eqn:p-vector-3}
{\bf g}=(g_0, g_1, \cdots, g_{2(N-1)})^T \in R^{2N-1},
\end{equation}
where $g_0=p_0$ and
\begin{eqnarray}\label{eqn:projection-3}
& g_{2k-1}=p_{2k-1} \mbox{ and } g_{2k}=0, & \mbox{ if } p_{2k-1} > 0, \\
& g_{2k-1}=0 \mbox{ and } g_{2k}=p_{2k},   & \mbox{ if } p_{2k} > 0.
\end{eqnarray}
for $k=1, \cdots, N-1$.
\end{enumerate}

\begin{figure}[htb]
\centering
\includegraphics[width=10cm]{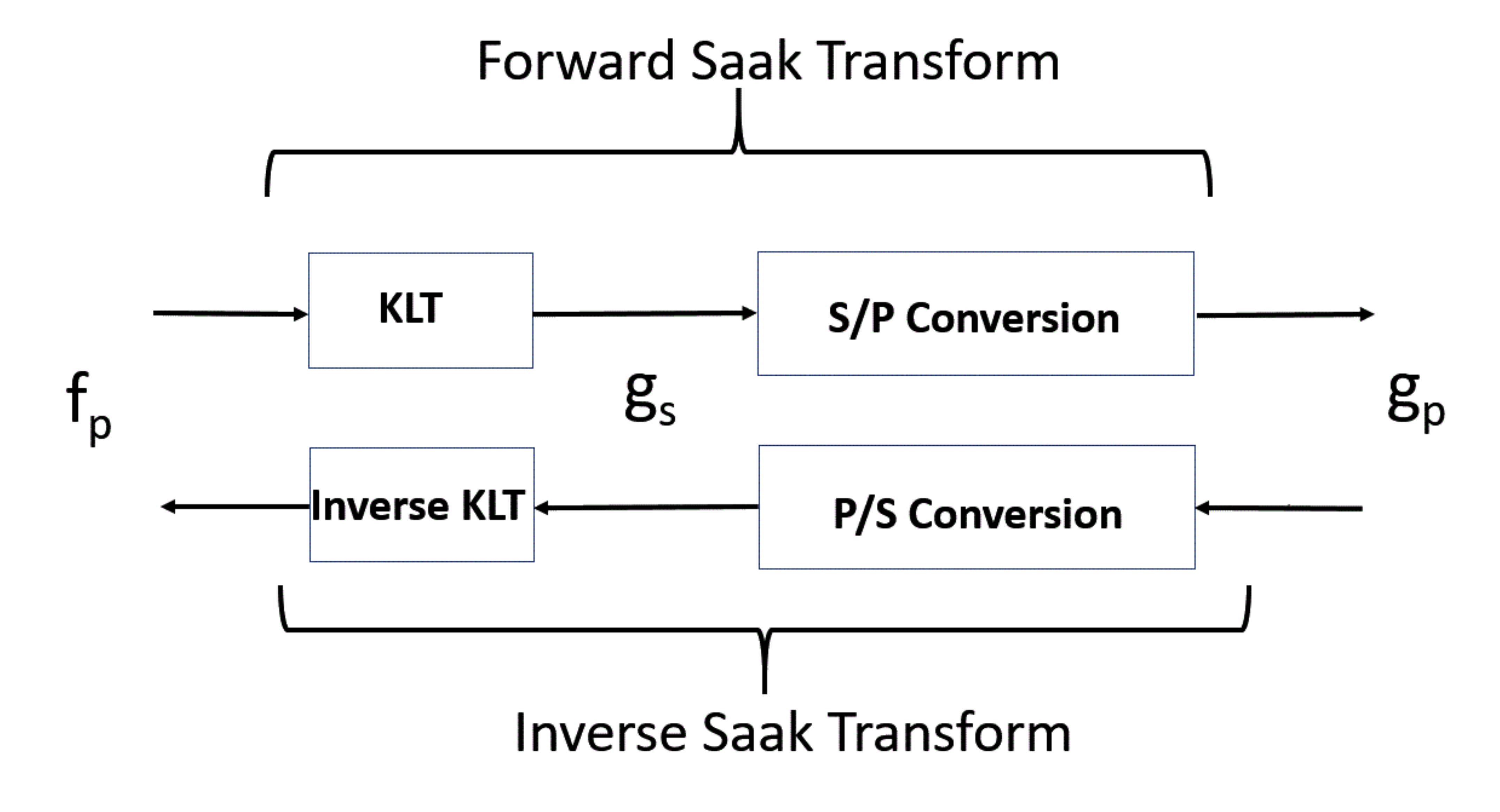} 
\caption{The block diagram of forward and inverse Saak transforms,
where ${\bf f}_p$, ${\bf g}_s$ and ${\bf g}_p$ are the input in
position format, the output in sign format and the output in position
format, respectively.}\label{fig:saak}
\end{figure}

The kernel augmentation scheme was described above in a way to motivate
the Saak transform. In practical implementation, we can offer another
view to the cascade of kernel augmentation and ReLU. As shown in Fig.
\ref{fig:aug_components}, projection values $p_k$ on KLT basis functions
${\bf b}_k$, $k=0, \cdots, N-1$ can be positive or negative.  It is
called the sign format of the projection output and denoted by
\begin{equation}\label{eqn:sign}
{\bf g}_s=(g_{s,0}, g_{s,1}, \cdots, g_{s,N-1} )^T \in R^{N},
\end{equation}
where 
\begin{equation}\label{eqn:project-21}
g_{s,k}= {\bf b}_k^T {\bf f} , \quad k=0, 1, \cdots, N-1.
\end{equation}
In contrast, the position of each AC element in Eq.
(\ref{eqn:p-vector-3}) is split into two consecutive positions.  Its
magnitude is recorded in the first and second positions, respectively,
depending on whether it is positive or negative. This is called the
position format of the projected output.  The cascade of kernel
augmentation and ReLU is equivalent to the sign-to-position (S/P) format
conversion as shown in Fig.  \ref{fig:saak}.  To simplify the
presentation below, we apply the S/P format conversion to the DC
component as well. Thus, the dimension of position format ${\bf g}_p$ is
twice of that of sign format ${\bf g}_s$.  To give an example, the
position format of $(5,-3)^T$ is $(5, 0, 0, 3)^T$, and vice versa. Note
that we demand the position-to-sign (P/S) format conversion for the
inverse Saak transform. 

\subsection{Inverse Saak Transform} 

The inverse Saak transform can be written as
\begin{equation}\label{eqn:subspace}
{\bf f}_p = \sum_{k=0}^{{N-1}} g_{s,k}{\bf b}_k,
\end{equation}
where ${\bf b}_k$ is a basis function. Given position format ${\bf
g}_p$, we need to perform the position-to-sign (P/S) format conversion
to get sign format ${\bf g}_s$ and then feed it to the inverse KLT.
This is illustrated in the lower branch of Fig. \ref{fig:saak}.

It is straightforward to verify the following.
\begin{itemize}
\item The input and the output of the forward/inverse KLT have the same $l_2$ distance.
\item The input and the output of the S/P format conversion and the P/S format
conversion have the same $l_1$ distance.
\end{itemize}
Besides, the $l_1$ distance is always no less than the $l_2$ distance.
Let ${\bf g}_{p,1}$ and ${\bf g}_{p,2}$ be two output vectors of input
vectors ${\bf f}_{p,1}$ and ${\bf f}_{p,2}$, respectively.  It is easy
to verify that
\begin{equation}\label{eqn:distance-1}
||{\bf g}_{p,1} - {\bf g}_{p,2}||_2 \le ||{\bf g}_{s,1} - {\bf g}_{s,2}||_2
= ||{\bf f}_{p,1} - {\bf f}_{p,2}||_2.
\end{equation}
As derived above, the $l_2$-distance of any two outputs of the forward
Saak transform is bounded above by the $l_2$-distance of their
corresponding inputs. This bounded-output property is important when
we use the outputs of the Saak transform (i.e. Saak coefficients) as the
features.  Furthermore, we have
\begin{equation}\label{eqn:distance-2}
||{\bf f}_{p,1} - {\bf f}_{p,2}||_2 = ||{\bf g}_{s,1} - {\bf g}_{s,2}||_2 
\le ||{\bf g}_{s,1} - {\bf g}_{s,2}||_1 = ||{\bf g}_{p,1} - {\bf g}_{p,2}||_1,
\end{equation}
which is important for the inverse Saak transform since the roles of
inputs and outputs are reversed. In this context, the $l_2$-distance of
any two outputs of the inverse Saak transform is bounded above by the
$l_1$-distance of their corresponding inputs. 

There are similarities and differences between the RECOS and the Saak
transforms. For similarities, both project the input onto each kernel
vector in the kernel set and adopt the ReLU to avoid sign confusion when
multiple linear transforms are in cascade.  For differences, the RECOS
transform does not specify the kernel number, which is determined
heuristically. Its kernel weights are optimized by backpropagation.  In
developing the Saak transform, we are concerned with efficient forward
and inverse transform simultaneously so that kernel orthonormality is
demanded. We argument transform kernels to annihilate the loss caused by
the ReLU.  The kernel number for a lossless Saak transform is uniquely
specified.  The approximation loss and the rectification loss of the
RECOS transform are completely eliminated in the lossless Saak
transform. 

\section{Multi-Stage Saak Transforms}\label{sec:M-Saak}

\begin{figure}[htb]
\centering
\includegraphics[width=8cm]{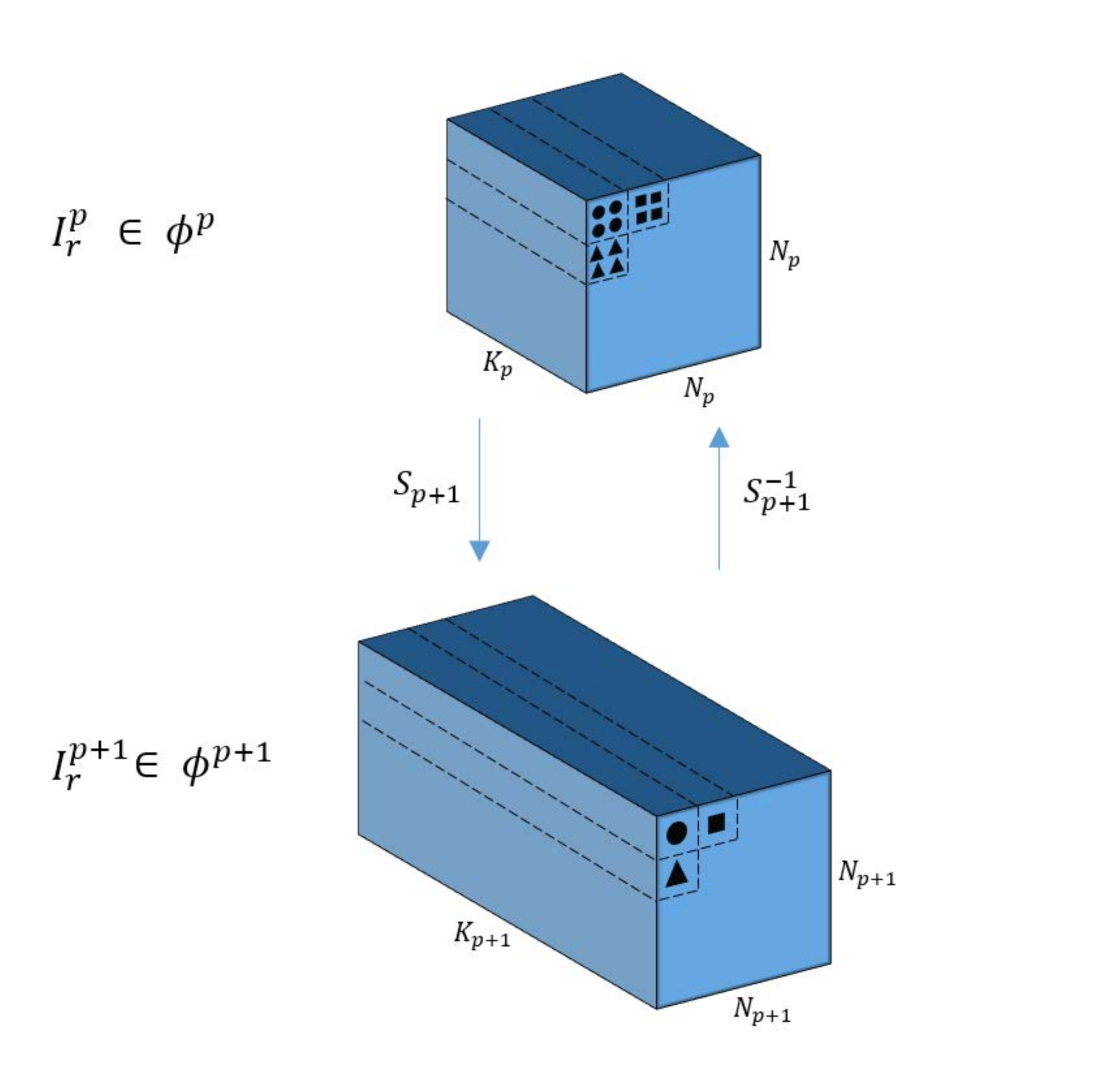} 
\caption{The relationship between local cuboids (LCs) and global
cuboids (GCs) in stages $p$ and $p+1$, where three representative LCs
are shown with solid dots, triangles and squares, respectively. The same
Saak transform is applied to all LCs and each of them yields one output
spectral vector. Sets $\Phi^p$ and $\Phi^{(p+1)}$ are collections of all
GCs in Stages $p$ and $(p+1)$ and images $I_r^p$ and $I_r^{(p+1)}$
are samples in $\Phi^p$ and $\Phi^{(p+1)}$, respectively.}\label{fig:Saak-2}
\end{figure}

\begin{figure}[htb]
\centering
\includegraphics[width=8cm]{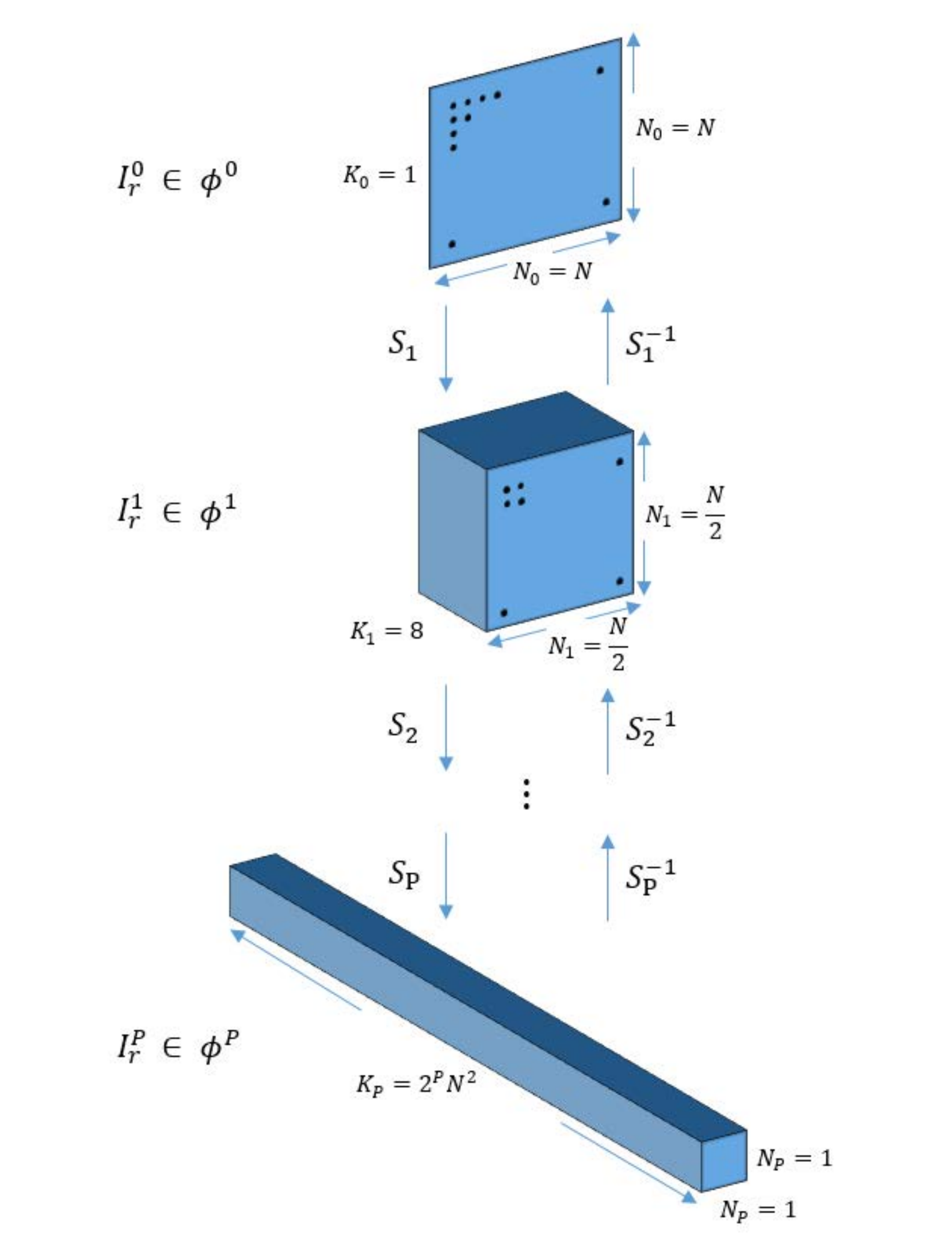} 
\caption{The conversion between an input image and its $P$
spatial-spectral representations via forward and inverse multi-stage
Saak transforms, where set $\Phi^p$ contains all GCs in the $p$th stage,
and $S_p$ and $S_p^{-1}$ are the forward and inverse Saak transforms
between stages $(p-1)$ and $p$, respectively.}\label{fig:GC}
\end{figure}

To transform images of a larger size, we decompose an image into four
quadrants recursively to form a quad-tree structure with its root being
the full image and its leaf being a small patch of size $2 \times 2$.
The first-stage Saak transform is applied at the leaf node.  Then,
multi-stage Saak transforms are applied from all leaf nodes to their
parents stage by stage until the root node is reached.  This process
can be elaborated below. 
\begin{itemize}
\item Stage 1: \\
Conduct the KLT on function values defined on
non-overlapping local cuboids (LCs) of size $2 \times 2 \times K_0$,
where $K_0=1$ for a monochrome image and $K_0=3$ for a color image, each
of which yields a set of signed KLT coefficients. The spatial dimension
of the entire input or the global cuboid (GC) is reduced by one half in
both horizontal and vertical directions.  
\item Stage $p=2, 3, \cdots$,
\begin{itemize}
\item Step (a): The signed KLT coefficients are converted to the
positioned KLT coefficients. The spectral dimension is doubled if
DC and AC components are treated in the same manner. 
\item Step (b): Conduct the KLT on positioned KLT coefficients defined
on non-overlapping LCs of dimension $(2 \times 2) \times 2 K_{p-1}$,
each of which yields a vector of signed KLT coefficients of dimension
$K_p=8 \times K_{p-1}$ as shown in Fig.  \ref{fig:Saak-2}.  Since the
spatial dimension of the input LC is $2 \times 2$ and the spatial
dimension of the output LC is $1\times 1$, the spatial dimension of the
GC is reduced by one half in both horizontal and vertical directions.
There is no explicit spatial pooling in multi-stage Saak transforms 
since non-overlapping cuboids are used here. 
\end{itemize}
\item Stopping Criterion \\
The whole process is stopped when we reach the last stage that has one set 
of signed KLT coefficients of dimension $1 \times 1 \times K_{f}$. 
If the image size is of $2^P \times 2^P$, we have $K_f = 2^{3P}$.
\end{itemize}
The signed KLT coefficients in the final stage are called the last-stage
Saak coefficients. The process is illustrated in Fig. \ref{fig:GC}.

The Saak transform is a mapping from a real-valued function defined on
cuboid $C(i_p, j_p, L_i, L_j, L_k)$ to an output vector. In practice,
its spatial dimensions $L_i$ and $L_j$ should be relatively small so
that the total number of grid points is under control. Here, we set
$L_i=L_j=2$. Hence, the Saak transform handles a cuboid of spatial
dimension $2\times 2$, which is called the local cuboid (LC).  In
contrast, the cuboid in set $\Phi^p$ as shown in Fig.  \ref{fig:GC} is
the GC in the $p$th stage and denoted by $GC_p$. A GC can be decomposed
into non-overlapping LCs.  The relationship between the $LC_p$ and the
$GC_p$ is illustrated in Fig. \ref{fig:Saak-2}.  The spatial dimension,
$N_p \times N_p$, of the input GC in Step (a) of Stage $(p+1)$, can be
recursively computed as
\begin{equation}\label{eqn:GC}
N_p = N_{p-1} 2^{-1}, \quad p=1, 2, \cdots \mbox{  and  }  N_0=N.
\end{equation}
It implies that $N_p=N \times 2^{-p}$. Since $N=2^P$, we have $N_P=1$. 

The $GC_p$ and $LCs$ have the same number of spectral components in each
stage. By augmenting DC and AC transform kernels equally, the
spatial-spectral dimension of $LC_p$ is $2 \times 2 \times K_p$, where
$K_p$ can be recursively computed as
\begin{equation}\label{eqn:LC}
K_p= 2 \times 4 \times K_{(p-1)}, \quad K_0=1, \quad p=1,2, \cdots
\end{equation}
The right-hand-side (RHS) of the above equation is the product of three
terms. The first one is due to the S/P conversion.  The second and third
terms are because that the degree of freedom of the input cuboid and the
output vector should be preserved through the KLT. 

With this expression, we can give a physical meaning to the Saak
transform from the input LC of dimension $2\times2\times K_{p-1}$ in
position format to the output LC of dimension $1\times1\times K_{p}$ in
sign format (i.e. before kernel augmentation). The forward Saak
transform merges the spectra of 4 child nodes (corresponding to 4
spatial sub-regions) into the spectrum of their parent node
(corresponding to the union of the 4 spatial sub-regions).  The inverse
Saak transform splits the spectrum of a parent node into the spectra of
its 4 child nodes.  In other words, the forward Saak transform is a
process that converts spatial variation to spectral variation while the
inverse Saak transform is a process that converts spectral variation to
spatial variation. Multi-stage Saak transforms provide a family of
spatial-spectral representations through recursion.  The spatial
resolution is the highest in the source image and the spectral
resolution is the highest in the output of the last stage Saak
transform.  The intermediate stages provide different spatial-spectral
trade offs.  

It is worthwhile to point out that we can adopt a different
decomposition to make the hierarchical tree shallower, say, splitting a
parent node into $m \times m$ ($m=3, 4, \cdots$ child nodes.  Then, the
tree has a depth of $\log_m N$.  As one traverses the tree from the leaf
to the root, the coverage (or the receptive field) is larger, the
spatial resolution of the GC is lower, and the spectral component number
is larger.  As a consequence of Eq.  (\ref{eqn:LC}), we have $K_p=8^p$,
which grows very quickly. In practice, we should leverage the energy
compaction property of the KLT and adopt the lossy Saak transform by
replacing the KLT with the truncated KLT (or PCA). We focus on the
lossless Saak transform due to space limitation in this work, and will
discuss the topic of lossy Saak transforms systematically and separately
in the future. 

Based on the above discussion, Saak transform kernels actually do not
depend on the geometrical layout of input elements but their statistical
correlation property. Thus, {\em if we can decompose a set of
high-dimensional random vectors into lower-dimensional sub-vectors
hierarchically, we can apply the Saak transform to these sub-vectors
stage by stage from bottom to top until the full random vector is
reached}. We can generalize multi-stage Saak transforms to other types
of signals, say, communication signals received by receiver arrays of an
irregular geometry that contain both spatial and temporal data as input
elements. 

\section{MNIST Image Reconstruction, Feature Extraction and Classification}
\label{sec:applications}

We use the MNIST dataset\footnote{http://yann.lecun.com/exdb/mnist/} as
an example to illustrate the image reconstruction, feature selection and
classification processes. Since we are only concerned with the signed
Saak coefficients in this section, we simply call them Saak coefficients
for convenience. 

\subsection{Distributions of Saak Coefficients}\label{sec:distribution}

\begin{figure}[!th]
\centering
\begin{subfigure}[b] {\linewidth}
\centering
\includegraphics[width=0.99\linewidth]{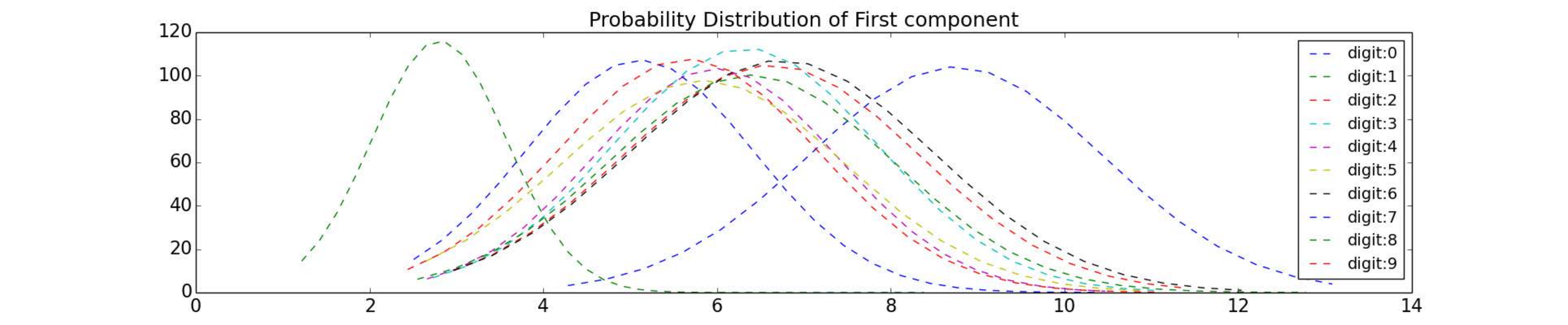}
\caption{The first spectral component}
\end{subfigure}
\centering
\begin{subfigure}[b]{\linewidth}
\centering
\includegraphics[width=0.99\linewidth]{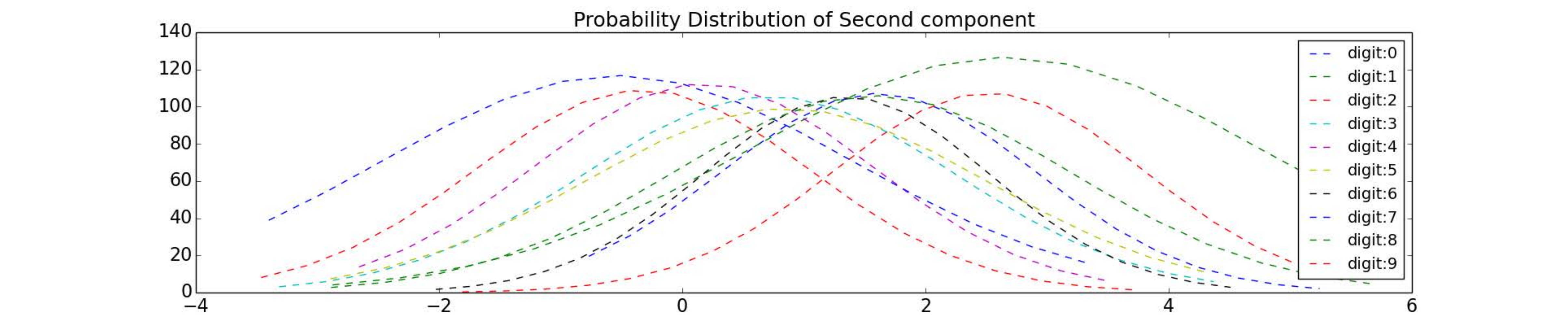}
\caption{The second spectral component}
\end{subfigure}
\centering
\begin{subfigure}[b]{\linewidth}
\centering
\includegraphics[width=0.99\linewidth]{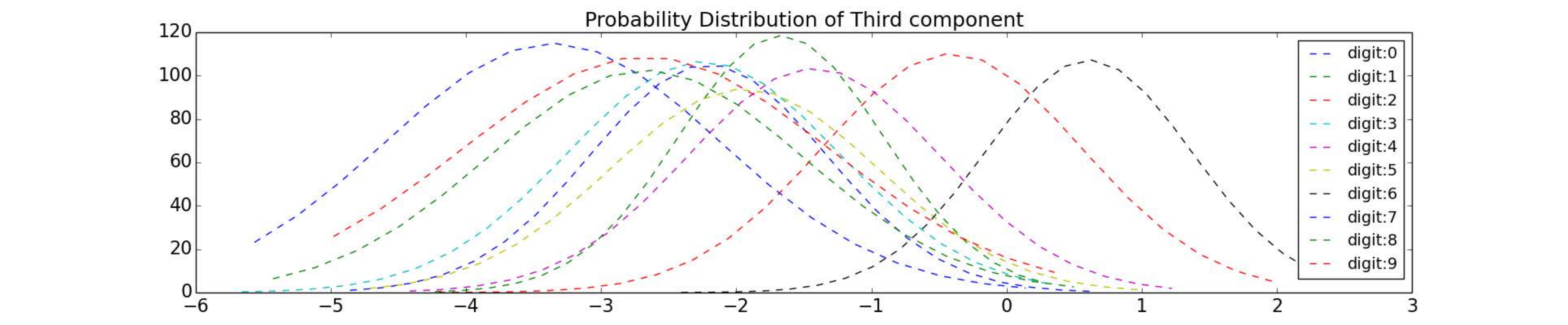}
\caption{The third spectral component}
\end{subfigure}
\caption{The distributions of Saak coefficients for the (a) first, 
(b) second and (c) third components for 10 digit classes.} 
\label{fig:Saak_distribution}
\end{figure}

There are 10 handwritten digit classes ranging from $0$ to $9$ in the
MNIST dataset. For image samples of the same digit, we expect their Saak
coefficients to have a clustering structure. We plot the distributions
of the first three last-stage Saak coefficients in Fig.
\ref{fig:Saak_distribution}.  Each plot has ten curves corresponding to
the smoothed histograms of ten digits obtained by 1000 image samples for
each digit. By inspection, each of them looks like a normal
distribution. To verify normality, we show the percentages of the
leading 256 last-stage Saak coefficients for each digit class that pass
the normality test with 100 and 1000 samples per class in Table
\ref{table:normality_test}. We adopt the Jarque-Bera test
\cite{jarque1987test} for the normality test and the Gubbbs' test
\cite{grubbs1950sample} for outlier removal in this table. 

\begin{table}[htbp!]
\normalsize
\centering
\caption{The percentages of the leading 256 last-stage Saak coefficients
that pass the normality test for each digit class before and after
outlier removal with 100 and 1000 samples per digit class, where S
denotes the number of samples.}\label{table:normality_test}
\begin{tabular}{|c|c|c|c|c|c|c|c|c|c|c|}\hline
Digit Class   & 0 & 1 & 2 & 3 & 4 & 5 & 6 & 7 & 8 & 9 \\ \hline \hline
S=100, raw & 91 & 73 & 96 & 90 & 87 & 89 & 88 & 88 & 90 & 86  \\ \hline
S=100, outlier removed & 97 & 89 & 98 & 97 & 97 & 96 & 96 & 93 & 98 & 96  \\ \hline
S=1000, raw & 63 & 30 & 74 & 68 & 56 & 66 & 53 & 46 & 61 & 53  \\ \hline
S=1000, outlier removed & 77 & 52 & 87 & 85 & 80 & 84 & 72 & 75 & 85 & 74  \\ \hline
\end{tabular}
\end{table}

We see from this table that, when each class has 100 samples, 90\% or
higher of these 256 last-stage Saak coefficients pass the normality test
after outlier removal. However, the percentages drop when the sample
number increases from 100 to 1000.  This sheds light on the difference
between small and big datasets. It is a common perception that the
Gaussian mixture model (GMM) often provides a good signal model.  Even
it is valid when the sample number is small, there is no guarantee that
it is still an adequate model when the sample number becomes very large.

\subsection{Image Synthesis via Inverse Saak transform}

\begin{figure}[htbp]
\centering
\includegraphics[width=12cm]{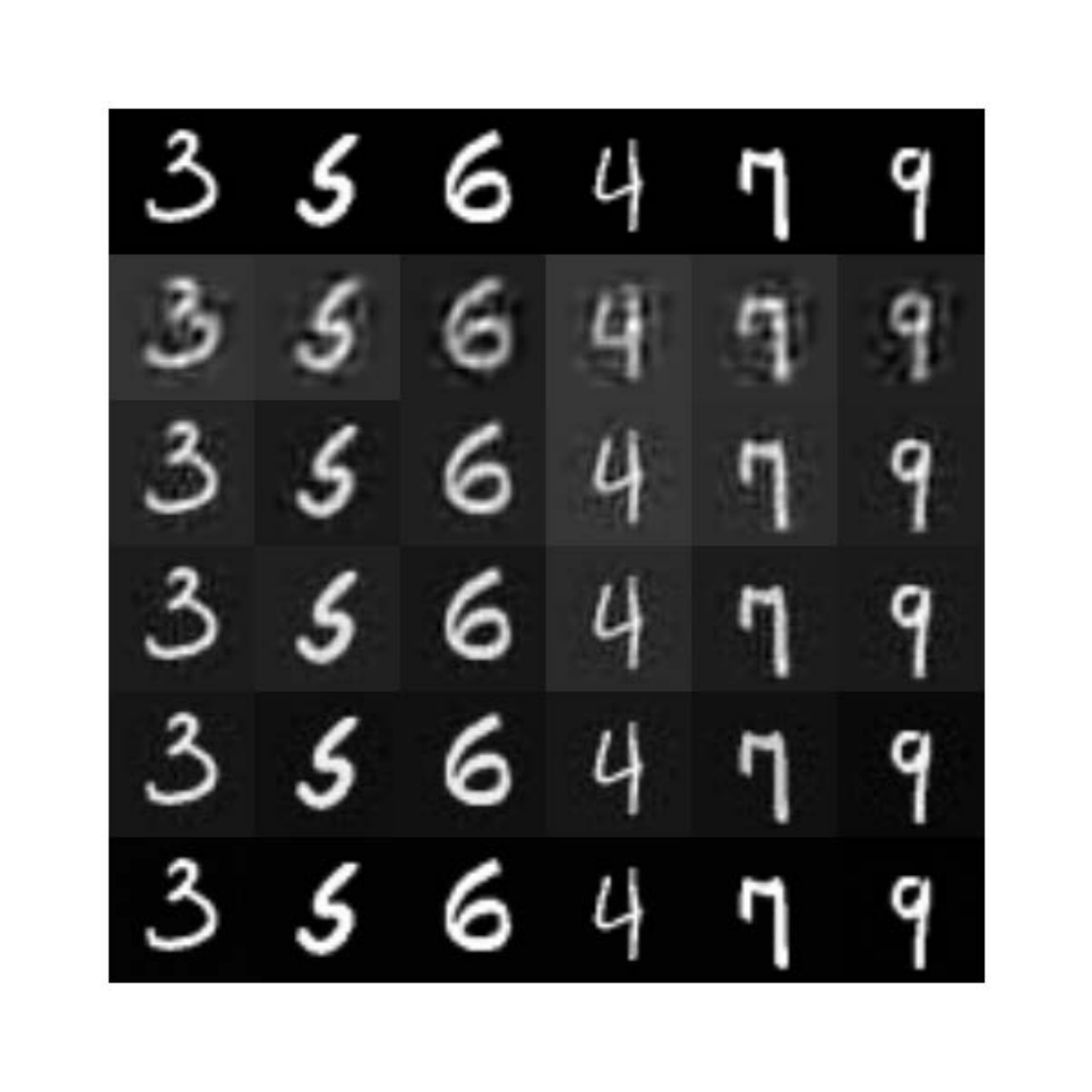}
\caption{Illustration of image synthesis with multi-stage inverse Saak
transforms (from top to bottom): six input images are shown in the first
row and reconstructed images based on the leading 100, 500, 1000 and 2000 
last-stage Saak coefficients are shown from the second to the fifth
rows. Finally, reconstructed images using all last-stage Saak
coefficients are shown in the last row. Images of the first and the last
rows are identical, indicating lossless full reconstruction.} \label{fig:synthesis}
\end{figure}

We can synthesize an input image with multi-stage inverse Saak
transforms with a subset of its last-stage Saak Coefficients. An example
is shown in Fig.  \ref{fig:synthesis}, where six input images are shown
in the first row and their reconstructed images using leading 100, 500,
1000, 2000 last-stage Saak coefficients are shown in the 2nd, 3rd, 4th
and 5th rows, respectively.  The last row gives reconstructed images
with all Saak coefficients (a total of $16,000$ coefficients). Images in
the first and the last rows are actually identical. It proves that
multi-stage forward and inverse Saak transforms are exactly the inverse
of each other numerically. The quality of reconstructed images with 1000
Saak coefficients is already very good. It allows people to see them
clearly without confusion.  Thus, we expect to get high classification
accuracy with these coefficients, which will be confirmed in Sec.
\ref{sec:results}.  Furthermore, it indicates that we can consider lossy
Saak transforms in all stages to reduce the storage and the
computational complexities. 

\subsection{Feature Selection}\label{sec:modeling}

The total number of Saak coefficients in lossless multi-stage Saak
transforms is large. To use Saak coefficients as image features
for classification, we can select a small subset of coefficients by
taking their discriminant capability into account.  To search Saak
coefficients of higher discriminant power, we adopt the F-test statistic
(or score) in the ANalysis Of VAriance (ANOVA), which is in form of
\begin{equation}\label{eqn:F-test}
F=\frac{\mbox{between-group variability (BGV)}}{\mbox{within-group variability (WGB)}}.
\end{equation}
The between-group variability (BGV) and the within-group variability 
(WGV) can be written, respectively, as
\begin{equation}\label{eqn:BGV}
\mbox{BGV}= \sum_{c=1}^C n_c (\bar{S}_c - \bar{S})^2/(C-1),
\end{equation}
and
\begin{equation}\label{eqn:WGV}
\mbox{WGV}= \sum_{c=1}^C \sum_{d=1}^{n_c} (S_{c,d} - \bar{S}_c)^2 / (T-C),
\end{equation}
where $C$ is the class number, $n_c$ is the number of Saak coefficients
of the $c$th class, $\bar{S}_c$ is the sample mean of the $c$th class,
$\bar{S}$ is the mean of all samples, $S_{c,d}$ is the $d$th sample of
the $c$th class, and $T$ is the total number of samples. If 6000 samples
per class is used in the training, we have $C=10$, $n_c=6,000$ and
$T=60,000$.  We select Saak coefficients of higher discriminant power by
computing their F-test scores. The larger the F-test score, the higher
the discriminant power. These selected Saak coefficients form the raw
feature vector. Then, we apply the dimension reduction technique further
to obtain the reduced-dimension feature vector. 

\subsection{MNIST Image Classification Results}\label{sec:results}

We use 60,000 image samples (namely, 6000 training images for each
digit) to compute the F-test score of all Saak coefficients. Then, we
test with the following three settings. 
\begin{itemize}
\item Setting No. 1: Selecting the leading 2000 last-stage Saak coefficients
(without considering their F-test scores). 
\item Setting No. 2: Selecting 2000 last-stage Saak coefficients with the 
highest F-test scores. 
\item Setting No. 3: Selecting 2000 Saak coefficients with the highest F-test 
scores from all stages. 
\end{itemize}
Furthermore, we conduct the PCA on these 2000 selected Saak coefficients
(called the raw feature vector) to reduce the feature dimension from
2000 to 64, 128 or 256 (called the reduced-dimension feature vector).
Finally, we apply the SVM classifier to the reduced-dimension feature
vectors of each test image. The classification accuracy is given in
Table \ref{table:accuracy_4}.  It is no a surprise that Setting NO. 1 is
the worst while Setting No. 3 is the best. However, it is worthwhile to
emphasize that the performance gap between Setting No. 1 and No. 2 is
very small. It shows the energy compaction power of the Saak transform.
If we select features only from the last-stage Saak coefficients, it
appears to be fine to focus on leading coefficients. However, to boost
the classification performance more significantly, we need to consider
Saak coefficients from earlier stages since these Saak coefficients take
both spatial and spectral information into account. 

\begin{table}[htbp]
\normalsize
\centering
\caption{The classification accuracy using the SVM classifier, where
each column indicates the reduced feature dimension from 2000 Saak
coefficients. Each row indicates a particular setting in selecting the
Saak coefficients.}\label{table:accuracy_4}
\begin{tabular}{|c|c|c|c|}\hline
   	       & 64    & 128   & 256     \\ \hline 
Setting No. 1  & 97.22 & 97.08 & 96.82   \\  \hline
Setting No. 2  & 97.23 & 97.10 & 96.88   \\  \hline
Setting No. 3  & 98.46 & 98.50 & 98.31   \\  \hline
\end{tabular}
\end{table}

Since Setting No. 3 gives the best performance, we choose this setting
and vary the dimension of raw and reduced-dimension feature vectors. The
classification accuracy results are shown in Table
\ref{table:accuracy_2}. We see that these results do not vary much. The
accuracy is all between 98\% and 99\%. The best result in this table is
98.52\%. Furthermore, we still choose setting No. 3 but adopt a
different classifier, which is the KNN classifier with $K=5$.  The
classification accuracy results are shown in Table
\ref{table:accuracy_3}. Again, we see that these results do not vary
much.  The accuracy is all between 97\% and 98\%. The best result in
this table is 97.53\%. The SVM classifier is better than the KNN
classifier in this example. 

\begin{table}[htbp]
\normalsize
\centering
\caption{The classification accuracy using the SVM classifier under
Setting No. 3, where each column indicates the raw feature
dimension and each row indicates the reduced feature dimension.}
\label{table:accuracy_2}
\begin{tabular}{|c|c|c|c|c|c|}\hline
	 & 1000 & 2000 & 3000 & 4000 & 5000 \\ \hline
64       & 98.49 & 98.46 & 98.42 & 98.43 & 98.41  \\  \hline
128      & 98.46 & 98.50 & 98.52 & 98.51 & 98.52   \\  \hline
256      & 98.20 & 98.31 & 98.27 & 98.29 & 98.25    \\  \hline
\end{tabular}
\end{table}

\begin{table}[htbp]
\normalsize
\centering
\caption{The classification accuracy using the KNN classifier under
Setting No. 3, where each column indicates the raw feature
dimension and each row indicates the reduced feature dimension.}
\label{table:accuracy_3}
\begin{tabular}{|c|c|c|c|c|c|}\hline
     & 1000 & 2000 & 3000 & 4000 & 5000 \\ \hline 
64   & 97.53 & 97.52 & 97.50 & 97.52 & 97.45  \\  \hline
128  & 97.49 & 97.45 & 97.46 & 97.37 & 97.39   \\  \hline
256  & 97.50 & 97.42 & 97.39 & 97.41 & 97.39   \\  \hline
\end{tabular}
\end{table}

\section{CNN versus Saak Transform: Comparative Study}\label{sec:discussion}

CNNs and multi-stage Saak transforms share some similarities in overall
system design. For example, both adopt the convolution (or transform)
operations to generate responses which are followed by the ReLU
operation before being fed into the next layer (or stage).  However,
there are major differences between them.  We compare them in four
aspects below. 

\subsection{End-to-end Optimization versus Modular Design}

A CNN is built upon the choice of an end-to-end network architecture and
an optimization cost function. The filter weights are optimized through
backpropagation which is an iterative optimization procedure. The Saak
transform solution follows the traditional pattern recognition framework
that by partitioning the whole recognition task into two separate
modules; namely, the feature extraction module and the classification
module. 

If the number of object classes is fixed, the training and testing data
are reasonably correlated, and the cost function is suitably defined,
the end-to-end optimization approach taken by CNNs offers the
state-of-the-art performance. However, this end-to-end optimization
methodology has its own weaknesses: 1) robustness against perturbation,
2) scalability against the class number, 3) portability among different
datasets. The first one has been reported in several papers, e.g.
\cite{nguyen2015deep}, \cite{moosavi2016deepfool},
\cite{fawzi2017geometric}. For the second one, if we increase or
decrease the object class number by one for the ImageNet, all filter
weights of the whole network, which consists of the feature extraction
subnet and the decision subnet, have to be retrained.  It is
understandable that the increase or decrease of object classes will
affect the decision subnet.  However, it is against intuition that
filter weights of the feature extraction subnet are also affected by the
object class number. For the third one, it explains the need of a large
amount of research for domain adaption \cite{donahue2014decaf},
\cite{ganin2015unsupervised}.  The end-to-end objective optimization of
CNNs offers the best performance for a specific task under a specific
setting at the cost of robustness, flexibility and generalizability. 

In contrast with CNN's end-to-end optimization methodology, the modular
design and the Saak transform approach is expected to be more robust
against perturbation and less sensitive to the variation of object
classes. The small perturbation will not affect leading last-stage Saak
coefficients much due to the use of PCA. Kernels in earlier Saak
transform stages should not change much if their covariance matrices do
not change much. If this is the case, we can use the same network to
generate different Saak features for new object classes. More studies
along this line will be reported. 

\subsection{Generative Model versus Inverse Transform}

No inverse transform has been studied for CNNs except the inverse RECOS
presented in Sec. \ref{sec:RECOS} of this work. On the other hand, two
approaches have been examined to generate images based on CNNs. One is
the Generative Adversarial Network (GAN) \cite{goodfellow2014generative}
and the other is the Variational AutoEncoder (VAE)
\cite{kingma2013auto}.  One has to train a generative network and a
discriminative network for the GAN solution, and an encoder network and
a decoder network for the VAE solution. Once the whole network system is
trained, one can use a latent vector to generate images through the
generative network and the decoder for GANs and VAEs, respectively. 

Here, we follow the traditional signal analysis and synthesis framework.
The kernels of the Saak transform are orthonormal, and the inverse Saak
transform can be easily defined and conducted conveniently.  For an
arbitrary input, the cascade of the forward and inverse multi-stage Saak
transforms results in an identity operator. This is a trivial case. It
is however possible to consider non-trivial applications. To take the
single image super-resolution problem as an example, we can build two
forward/inverse Saak transform pipes - one for the low resolution images
and the other for the high resolution images. We may try the following
high-level idea. That is, one can build bridges between these two pipes
and, then, feed the low-resolution images into the low-resolution
forward transform pipe and get the high-resolution images from the
high-resolution inverse transform pipe.  This is an interesting research
topic for further exploration. 

\subsection{Theoretical Foundation}

Significant efforts have been made to shed light on the superior
performance of CNNs. The early work of Cybenko
\cite{cybenko1989approximation} and Hornik {\em et al.}
\cite{hornik1989multilayer} interpreted the multi-layer perceptron (MLP)
as a universal approximator.  Recent theoretical studies on CNNs include
scattering networks \cite{mallat2012group, bruna2013invariant,
wiatowski2015mathematical}, tensor analysis \cite{cohen2015expressive},
generative modeling \cite{dai2014generative}, relevance propagation
\cite{bach2015pixel}, Taylor decomposition
\cite{montavon2015explaining}, the multi-layer convolutional sparse
coding (ML-CSC) model \cite{sulam2017multi}, over-parameterized shallow
neural network optimization \cite{soltanolkotabi2017theoretical}, etc.
Another popular research topic is the visualization of filter responses
at various layers \cite{simonyan2013deep, zeiler2014visualizing,
zhou2014object}.  Despite intuition was given in the above-mentioned
work, complete CNN theory is lacking. As the complexity of
recent CNN architectures goes higher, the behavior of these networks is
mathematically intractable.  In contrast, The Saak transform is fully
built upon linear algebra and statistics. It can be easily and fully 
explained mathematically. 

\subsection{Filter Weight Determination, Feature Selection and Others}

There is a fundamental difference in filter weight determination between
CNNs and Saak-transform-based machine learning methodology.  Filter
weights in CNNs are determined by data and their associated labels.
They are updated by backpropagation using such information pair provided
at two ends of the network.  A CNN learns its best parameters by
optimizing an objective function iteratively. The iteration number is
typically very large. We propose another fully automatic filter weight
selection scheme.  The Saak transform select its kernel functions (or
filter weights) using the KLT stage by stage. It is built upon the
second-order statistics of the input data. Neither data labels nor
backpropagation is needed.  Data labels are only needed in the decision
module, which is decoupled from the Saak transform module. 

Before the resurgence of CNNs, feature extraction was probably the most
important ingredient in pattern recognition. Different applications
demanded different features, and they were extracted heuristically.
These features are called the ``handcrafted" features nowadays. One of
the key characteristics of the CNN approach is that features can be
``learned" automatically from the data and their labels through
backpropagation. These iteratively optimized features are called the
``deep" or the ``learned" features. Here, we provide a third approach to
feature extraction based on multi-stage Saak transforms. They are the
Saak features. To obtain Saak features, we need data labels but use them
in a different manner.  Multi-stage Saak coefficients of samples from
the same class are computed to build feature vectors of that class.  We
should collect those Saak coefficients that have higher discriminant
power among object classes to reduce the feature dimension. 

There are differences in implementation details. CNNs conduct
convolutional operations on overlapping blocks/cuboids and adopt spatial
pooling to reduce the spatial dimension of responses before they are fed
into the next layer. The Saak transform is applied to non-overlapping
blocks/cuboids and no spatially pooling is needed. It achieves spatial
dimension reduction with no additional effort.  The horizontal and
vertical spatial dimensions of CNN filters typically take odd numbers,
say $3 \times 3$, $5 \times 5$, $11 \times 11$, etc. The spatial
dimension of the Saak transform in each stage can be even or odd. The
CNN parameter setting mostly follows prior art as a convention.  The
parameters of the Saak transform can be determined with theoretical
support. 

\section{Conclusion and Future Work}\label{sec:conclusion}

Data-driven forward and inverse Saak transforms were proposed. By
applying multi-stage Saak transforms to a set of images, we can derive
multiple representations of these images ranging from the pure spatial
domain to the pure spectral domain as the two extremes.  There is a
family of joint spatial-spectral representations with different
spatial-spectral trade offs between them.  The Saak transform offers a
new angle to look at the image representation problem and provides
powerful spatial-spectral Saak features for pattern recognition. The
MNIST dataset was used as an example to demonstrate this application. 

There are several possible extensions. First, we focused on the lossless
Saak transform in this work. It is desired to study the lossy Saak
transform by removing spectral components of less significance in a
systematic fashion. Second, we used the F-test score to select Saak
coefficients of higher discriminant power. There may be other more
effective methods to extract discriminant features from a huge number of
Saak coefficients to achieve better classification performance.  Third,
the Saak transform can be applied to any high dimensional random vectors
as long as they can be decomposed into lower dimensional sub-vectors
hierarchically. Along this line, we may generalize the
Saak-transform-based feature extraction method to other types of
signals, say, communication signals. Fourth, the inverse Saak transform
can be used for image synthesis/generation.  It is interesting to
investigate the Saak-transform-based image generation approach and
compare it with other approaches such as VAEs and GANs. 

\section*{Acknowledgment}

This material is partially based on research sponsored by DARPA and Air
Force Research Laboratory (AFRL) under agreement number
FA8750-16-2-0173. The U.S. Government is authorized to reproduce and
distribute reprints for Governmental purposes notwithstanding any
copyright notation thereon.  The views and conclusions contained herein
are those of the authors and should not be interpreted as necessarily
representing the official policies or endorsements, either expressed or
implied, of DARPA and Air Force Research Laboratory (AFRL) or the U.S.
Government. This project was also partially supported by the National
Heart Lung Institute (R01HL129727)(T.K.H.). 

\bibliographystyle{elsarticle-num} 
\bibliography{CNN}

\end{document}